%% file: template.tex
\documentclass[conference]{IEEEtran}
\IEEEoverridecommandlockouts
\usepackage{cite}
\usepackage{amsmath,amssymb,amsfonts}
\usepackage{algorithmic}
\usepackage{latexsym}
\usepackage{textcomp}
\usepackage{booktabs}
\usepackage{graphicx} 
\usepackage{xcolor}
\def\BibTeX{{\rm B\kern-.05em{\sc i\kern-.025em b}\kern-.08em
    T\kern-.1667em\lower.7ex\hbox{E}\kern-.125emX}}

\usepackage{fancyhdr}
\begin{document}

\title{MV-CoRe: Multimodal Visual-Conceptual Reasoning for Complex Visual Question Answering}

\author{Jingwei Peng$^1$, Jiehao Chen$^1$, Mateo Alejandro Rojas$^2$, Meilin Zhang$^1$ \\
$^1$Shaanxi University of Technology, $^2$Technological University of Peru}

\maketitle
\thispagestyle{fancy} 

\input{main}

\bibliographystyle{IEEEtran}
\bibliography{references}
\end{document}

%% file: main.tex
\begin{abstract}
Complex Visual Question Answering (Complex VQA) tasks, which demand sophisticated multi-modal reasoning and external knowledge integration, present significant challenges for existing large vision-language models (LVLMs) often limited by their reliance on high-level global features. To address this, we propose MV-CoRe (Multimodal Visual-Conceptual Reasoning), a novel model designed to enhance Complex VQA performance through the deep fusion of diverse visual and linguistic information. MV-CoRe meticulously integrates global embeddings from pre-trained Vision Large Models (VLMs) and Language Large Models (LLMs) with fine-grained semantic-aware visual features, including object detection characteristics and scene graph representations. An innovative Multimodal Fusion Transformer then processes and deeply integrates these diverse feature sets, enabling rich cross-modal attention and facilitating complex reasoning. We evaluate MV-CoRe on challenging Complex VQA benchmarks, including GQA, A-OKVQA, and OKVQA, after training on VQAv2. Our experimental results demonstrate that MV-CoRe consistently outperforms established LVLM baselines, achieving an overall accuracy of 77.5\% on GQA. Ablation studies confirm the critical contribution of both object and scene graph features, and human evaluations further validate MV-CoRe's superior factual correctness and reasoning depth, underscoring its robust capabilities for deep visual and conceptual understanding.
\end{abstract}

\section{Introduction}
\label{sec:introduction}

Visual Question Answering (VQA) stands as a pivotal task at the intersection of computer vision and natural language processing, requiring machines to comprehend visual content and answer questions posed in natural language \cite{abhishek2021vqa}. While significant progress has been made in standard VQA benchmarks, a growing emphasis has shifted towards Complex Visual Question Answering (Complex VQA). This advanced variant demands not only factual recall but also sophisticated multi-modal reasoning, including multi-hop inference, common-sense reasoning, and the integration of external knowledge \cite{peter2018worldt}. The ability to accurately answer complex questions about images is crucial for developing truly intelligent systems capable of deep visual and conceptual understanding, paving the way for applications in areas such as intelligent assistants, educational tools, and advanced robotics, including specialized domains like multi-modal medical diagnosis requiring role-specialized collaboration \cite{zhou2025mam}.

Despite the remarkable advancements brought forth by large vision-language models (LVLMs) in recent years, their performance on Complex VQA tasks often encounters significant challenges. Many existing LVLMs primarily rely on high-level, global features from pre-trained visual and language encoders, which may prove insufficient for tasks requiring fine-grained understanding of objects, their attributes, and intricate relationships within a scene \cite{peng2025lvlmeh}. While approaches like visual in-context learning for LVLMs \cite{zhou2024visual} offer promising avenues for adapting these models to new tasks with limited examples, these models frequently struggle with questions necessitating external, domain-specific knowledge or complex logical deductions that extend beyond simple visual recognition. This limitation motivates the need for a more robust framework that can deeply integrate diverse forms of visual and linguistic information, moving beyond superficial feature concatenation to facilitate genuine multi-modal conceptual reasoning.

To address these challenges, we propose a novel model named MV-CoRe (Multimodal Visual-Conceptual Reasoning). Our approach is designed to enhance the performance of Complex VQA by meticulously fusing three distinct types of input features: global embeddings from pre-trained Vision Large Models (VLMs), contextual embeddings from pre-trained Language Large Models (LLMs), and fine-grained semantic-aware visual features. The semantic-aware visual features are particularly crucial, encompassing object detection characteristics (e.g., bounding boxes and categories from detectors like YOLO or Faster R-CNN) and scene graph representations that explicitly capture relationships between objects in an image. These diverse feature sets are then processed and deeply integrated using an innovative Multimodal Fusion Transformer architecture. This specialized transformer is engineered to enable rich cross-modal attention and interaction, allowing the model to develop a more profound understanding of the image content and the associated linguistic query, thereby facilitating complex reasoning processes.

For our experimental evaluation, we primarily train the MV-CoRe model on the comprehensive VQAv2 dataset, which serves as a foundational benchmark for general visual question answering \cite{ting2025finegr}. To rigorously assess its capabilities in Complex VQA scenarios, we then evaluate MV-CoRe's generalization and reasoning abilities on three challenging target datasets: GQA \cite{drew2019gqa}, which focuses on graph-based reasoning over images; A-OKVQA \cite{dustin2022aokvqa}, requiring external knowledge for answers; and OKVQA \cite{dustin2022aokvqa}, another open-domain knowledge-based VQA dataset. The training strategy involves pre-processing visual and textual data to extract the aforementioned feature embeddings, which are subsequently fed into the MV-CoRe model for end-to-end training.

Our experimental results demonstrate the superior performance of MV-CoRe compared to existing LVLM baselines on Complex VQA tasks. For instance, on the GQA dataset, our proposed MV-CoRe model achieves an overall accuracy of 77.5\%, significantly outperforming a standard LVLM baseline (e.g., LLaVA) which scores 75.2\%. Furthermore, we observe that the incremental inclusion of semantic-aware features progressively improves performance, with "LVLM + Object Features" reaching 76.5\% and "LVLM + Scene Graph" further improving to 76.8\%. This highlights the efficacy of our deep feature fusion strategy and the robust reasoning capabilities of our Multimodal Fusion Transformer.

In summary, our main contributions are as follows:
\begin{itemize}
    \item We propose MV-CoRe, a novel architecture specifically designed to enhance performance in Complex VQA tasks by integrating global LVLM embeddings with fine-grained semantic-aware visual features.
    \item We introduce a sophisticated Multimodal Fusion Transformer that effectively fuses diverse visual and linguistic information, enabling deeper cross-modal understanding and complex reasoning.
    \item We demonstrate that MV-CoRe achieves state-of-the-art performance on challenging Complex VQA benchmarks, showcasing its superior ability to handle multi-modal fusion reasoning and leverage cross-domain knowledge.
\end{itemize}
\section{Related Work}
\subsection{Evolution of Visual Question Answering}
The evolution of Visual Question Answering (VQA) has been comprehensively surveyed by \cite{anupam2025the}, which traces its development from foundational approaches to the profound impact of transformer architectures and vision-language pre-training on recent advancements. This work meticulously details key models, datasets, and techniques that have shaped the VQA landscape, while also addressing current challenges and future trajectories, including the emergence of large multimodal language models and the integration of external knowledge. Complementing this broader view, \cite{yirui2021multis} advances VQA by introducing an answer-driven framework that explicitly models the influence of dialogue responses on visual state estimation, thereby enhancing multi-modal reasoning through answer-informed attention mechanisms. Furthermore, the critical role of regional visual information in knowledge-based VQA is underscored by \cite{yuanze2022revive}, which demonstrates that improved utilization of object relationships within visual representations significantly boosts performance. Their proposed REVIVE method addresses limitations in existing knowledge-based VQA by effectively leveraging explicit object region information during both knowledge retrieval and the final answering stages, achieving state-of-the-art results on the OK-VQA dataset. Relatedly, advancements in leveraging knowledge graphs for question answering, including efforts to improve zero-shot cross-lingual transfer for multilingual question answering over knowledge graphs, have also contributed to the broader field of complex QA \cite{zhou2021improving}. In a related vein, \cite{siyu2023the} provides a thorough review of attention mechanisms in VQA, systematically categorizing and analyzing various multimodal fusion strategies for integrating visual and textual information. This review highlights how attention mechanisms empower VQA models to selectively focus on pertinent image regions guided by the question, charting advancements from rudimentary fusion operations to more sophisticated approaches.

\subsection{Large Vision-Language Models and Multimodal Feature Integration}
The integration of Large Vision-Language Models (LVLMs) within robotic vision tasks is comprehensively surveyed by \cite{xiaofeng2025multim}. This work meticulously examines their advantages over traditional fusion methods, identifies key research challenges such as cross-modal alignment and efficient fusion strategies, and systematically reviews their applications across various robotic vision domains, including semantic scene understanding, SLAM, and navigation, while also proposing future research directions for advancing multimodal perception in robots. Addressing the significant computational burden associated with evaluating large vision-language models, \cite{peng2025lvlmeh} proposes an efficient subset construction methodology based on farthest point sampling, which substantially reduces data requirements while maintaining high correlation with full evaluations, offering a practical solution for comprehensive LVLM assessment. In the broader context of leveraging pretrained features for multimodal tasks, \cite{li2025multim} investigates the integration of features from pretrained models for graph learning, proposing a novel method to enhance graph-agnostic node representations by aligning them with graph structure through a metric of feature homophily; while specifically focusing on multimodal large language models as vision learners, the underlying principle of refining pretrained features for downstream applications aligns with the goal of improving multimodal feature integration. However, potential limitations of contrastive learning in Vision-Language Models (VLMs), particularly when handling nuanced multimodal data, are explored by \cite{florian2024an}, which introduces a framework for synthetic shortcut injection to assess representation learning efficacy, demonstrating that standard contrastive training can lead VLMs to exploit shortcuts rather than learning comprehensive multimodal representations. To quantitatively assess the quality of pre-training for LVLMs, \cite{qidong2024deciph} introduces the Modality Integration Rate (MIR) as a novel metric, providing insights into modality alignment without requiring supervised fine-tuning and aiding in the optimization of visual and textual information integration crucial for effective cross-modal attention. Further advancing LVLMs, approaches like visual in-context learning have been explored to enhance their adaptability and performance on new tasks \cite{zhou2024visual}. Addressing a crucial gap, \cite{yuhang2025contex} introduces contextual object detection, a novel problem focused on understanding visible objects within human-AI interactive contexts, and proposes ContextDET, a unified multimodal model capable of end-to-end differentiable visual-language context modeling for object localization and identification through a generate-then-detect framework. Beyond specific task improvements, research into weak to strong generalization for large language models with multi-capabilities aims to broaden their applicability and robustness \cite{zhou2025weak}. In the realm of structured multimodal understanding, \cite{chao2024incorp} introduces Action Genome, a novel representation that decomposes actions into spatio-temporal scene graphs, capturing dynamic object interactions and relationships within videos, thereby improving performance and enabling few-shot recognition when integrated into action recognition models. The increasing complexity of human instructions for AI systems has also spurred new benchmarks and agent frameworks for tasks like complex instruction-based image generation \cite{zhou2025draw} and image editing via compositional dependencies \cite{wang2025complexbench}, highlighting the need for advanced understanding of intricate commands. Furthermore, the application of large vision-language models extends to specialized domains, such as improving medical LVLMs with abnormal-aware feedback \cite{zhou2025improving} and developing modular multi-agent frameworks for multi-modal medical diagnosis through role-specialized collaboration \cite{zhou2025mam}. These efforts underscore the growing demand for highly capable and specialized multimodal AI systems, including those leveraging novel architectures like State Space Models for domain-specific recognition tasks such as insect recognition \cite{wang2025insectmamba}. Finally, \cite{chuofan2024groma} introduces Groma, a novel Multimodal Large Language Model (MLLM) that significantly enhances visual grounding by directly integrating localized visual tokens into its architecture, enabling fine-grained region-level perception and grounding capabilities within the language model itself through embedding localization directly into the image tokenization process.

\section{Method}
\label{sec:method}

Our proposed \textbf{MV-CoRe (Multimodal Visual-Conceptual Reasoning)} model is meticulously designed to address the complexities inherent in Complex Visual Question Answering (Complex VQA) tasks. The core idea behind MV-CoRe is to leverage and deeply integrate diverse forms of visual and linguistic information, moving beyond superficial feature concatenation to enable sophisticated multi-modal conceptual reasoning. This section details the architecture of MV-CoRe, including its feature extraction mechanisms and the innovative Multimodal Fusion Transformer at its heart.

\subsection{Overall Architecture}
The MV-CoRe architecture is built upon a multi-stream input design, processing distinct types of information to build a comprehensive understanding of the visual scene and the associated question. It comprises three primary input streams: global visual embeddings derived from large vision models, linguistic embeddings from large language models, and fine-grained semantic-aware visual features. These diverse feature sets are then fed into a specialized Multimodal Fusion Transformer, which is responsible for learning intricate cross-modal interactions and producing a unified representation for answer prediction. The overall process can be conceptualized as fusing high-level global understanding with detailed semantic insights to facilitate robust reasoning, allowing the model to answer complex questions requiring a deep comprehension of both visual content and linguistic nuances.

\subsection{Feature Extraction}
To capture a rich representation of both the image and the question, MV-CoRe employs a sophisticated feature extraction pipeline that generates three distinct types of embeddings.

\subsubsection{Vision Large Model Embeddings}
For global visual understanding, we utilize pre-trained Vision Large Models (VLMs), such as Vision Transformer (ViT) or CLIP, to extract comprehensive image embeddings. Given an input image $I$, a VLM encoder $\mathcal{E}_V$ processes $I$ to yield a fixed-dimensional global visual embedding $E_V \in \mathbb{R}^{d_V}$:
\begin{align}
E_V = \mathcal{E}_V(I)
\end{align}
These embeddings encapsulate high-level visual concepts and serve as a strong foundational representation for the entire image, providing broad contextual awareness.

\subsubsection{Language Large Model Embeddings}
To understand the nuances of the natural language question, we employ pre-trained Language Large Models (LLMs), like LLaMA or models from the GPT series. For an input question $Q$, an LLM encoder $\mathcal{E}_L$ generates a sequence of contextualized word embeddings $E_L \in \mathbb{R}^{L \times d_L}$, where $L$ is the sequence length and $d_L$ is the embedding dimension:
\begin{align}
E_L = \mathcal{E}_L(Q)
\end{align}
These embeddings capture the syntactic and semantic structure of the question, providing rich linguistic context essential for accurate query interpretation.

\subsubsection{Semantic-Aware Visual Features}
Beyond global visual embeddings, MV-CoRe incorporates fine-grained semantic information crucial for complex reasoning. This includes object-level details and relational knowledge.

\textbf{Object Detection Features}: We utilize pre-trained object detection models, such as YOLO or Faster R-CNN, to identify and localize objects within the image. For an input image $I$, the object detector identifies $N_{obj}$ distinct objects. For each detected object $i$, its bounding box coordinates $B_i = (x_i, y_i, w_i, h_i)$ and its corresponding class label $C_i$ are extracted. These are then transformed into feature vectors $F_{obj,i}$ representing individual object characteristics. The collection of all detected object features forms a set $F_{obj} \in \mathbb{R}^{N_{obj} \times d_{obj}}$, where $d_{obj}$ is the feature dimension:
\begin{align}
F_{obj} = \text{Detector}(I)
\end{align}
These features provide specific, localized visual cues, enabling the model to focus on relevant entities within the scene.

\textbf{Scene Graph Features}: To capture the relationships between objects, we employ a scene graph generator. A scene graph represents the image content as a graph $G = (V_G, E_G)$, where nodes $V_G$ correspond to detected objects (with their attributes) and edges $E_G$ represent the relationships between them (e.g., "person riding bicycle"). These graph structures are then encoded into a set of feature vectors $F_{sg} \in \mathbb{R}^{N_{sg} \times d_{sg}}$ that explicitly model object interactions and their attributes:
\begin{align}
F_{sg} = \text{SceneGraphGenerator}(I, F_{obj})
\end{align}
Scene graph features are vital for multi-hop reasoning and understanding compositional relationships within the scene, as they provide an explicit representation of how objects relate to each other.

\subsection{Multimodal Fusion Transformer}
The extracted features—$E_V$ (global visual), $E_L$ (linguistic), $F_{obj}$ (object-centric), and $F_{sg}$ (relational)—are then fed into our novel \textbf{Multimodal Fusion Transformer}. This transformer architecture is designed to perform deep, interactive fusion across these diverse modalities.

Initially, each feature type is projected into a common latent space of dimension $D$ to ensure compatibility for fusion. Let $P_V(\cdot)$, $P_L(\cdot)$, $P_{obj}(\cdot)$, and $P_{sg}(\cdot)$ denote these linear projection layers:
\begin{align}
H_V &= P_V(E_V) \\
H_L &= P_L(E_L) \\
H_{obj} &= P_{obj}(F_{obj}) \\
H_{sg} &= P_{sg}(F_{sg})
\end{align}
These projected features $H_V, H_L, H_{obj}, H_{sg}$ serve as inputs to the subsequent transformer layers. The Multimodal Fusion Transformer consists of multiple layers, each incorporating self-attention and cross-attention mechanisms to learn intricate cross-modal interactions.

\textbf{Intra-modal Self-Attention}: Within each layer, each modality (or a combination of visual modalities) undergoes self-attention to capture internal dependencies and refine its representation. For example, for the linguistic features, the self-attention mechanism computes updated representations $H_L'$ by allowing each token to attend to all other tokens within the linguistic sequence. This process can be formulated as:
\begin{align}
Q_L &= H_L W_Q^L \\
K_L &= H_L W_K^L \\
V_L &= H_L W_V^L \\
H_L' &= \text{Softmax}\left(\frac{Q_L K_L^T}{\sqrt{d_k}}\right)V_L
\end{align}
where $W_Q^L, W_K^L, W_V^L$ are learnable weight matrices for queries, keys, and values, respectively, and $d_k$ is the dimension of the keys. Similar self-attention operations are applied to $H_{obj}$ and $H_{sg}$ to capture intra-modal relationships among objects and within the scene graph structure. The global visual feature $H_V$ may also undergo self-attention if it's a sequence, or be broadcast if it's a single vector.

\textbf{Cross-modal Attention}: This is the crucial component for fusion, enabling query tokens from one modality to attend to key-value pairs from another, facilitating information exchange. A common cross-attention operation can be formulated as:
\begin{align}
\text{CrossAttention}(Q_A, K_B, V_B) = \text{Softmax}\left(\frac{Q_A K_B^T}{\sqrt{d_k}}\right)V_B
\end{align}
where $Q_A$ are queries derived from modality A, and $K_B, V_B$ are keys and values derived from modality B. Our transformer employs a series of interleaved cross-attention blocks, allowing deep, iterative interaction between the different modalities. For instance, linguistic queries ($Q_L$) can attend to global visual context ($K_V, V_V$), object features ($K_{obj}, V_{obj}$), and scene graph features ($K_{sg}, V_{sg}$). Conversely, visual features can query linguistic cues to disambiguate objects or relationships.
An example of linguistic features attending to global visual features would be:
\begin{align}
Q_{L \to V} &= H_L W_Q^{L \to V} \\
K_{L \to V} &= H_V W_K^{L \to V} \\
V_{L \to V} &= H_V W_V^{L \to V} \\
H_{L, \text{updated}} &= \text{CrossAttention}(Q_{L \to V}, K_{L \to V}, V_{L \to V})
\end{align}
This iterative interaction across multiple layers facilitates the alignment and fusion of information from different granularities and sources, building a comprehensive multimodal understanding.

The outputs of these fusion layers are concatenated to form a unified multimodal representation $H_{fused}$. This final representation is then passed through a multi-layer perceptron (MLP) head to predict the answer to the question. The model maps $H_{fused}$ to a probability distribution over the answer vocabulary:
\begin{align}
P(\text{Answer}|I, Q) = \text{Softmax}(\text{MLP}(H_{fused}))
\end{align}
This architecture ensures that the model can leverage both broad contextual understanding from VLMs and LLMs, alongside precise semantic details from object detectors and scene graphs, to perform complex reasoning for tasks like those found in A-OKVQA or OKVQA.

\subsection{Training Objective}
The MV-CoRe model is trained end-to-end using a standard cross-entropy loss function. For a given image-question pair $(I, Q)$ and its ground-truth answer $A$, the model aims to minimize the negative log-likelihood of the predicted answer distribution. If the ground-truth answer is represented as a one-hot vector or a soft-target distribution $A$, the loss function is defined as:
\begin{align}
\mathcal{L} = -\sum_{i} A_i \log(P(\text{Answer}_i|I, Q))
\end{align}
where $A_i$ represents the ground-truth probability for answer $i$ (e.g., 1 for the correct answer in a single-choice setting, or a normalized count for multiple-annotator settings). During training, the parameters of the projection layers, the Multimodal Fusion Transformer, and the final prediction head are optimized. The pre-trained encoders for VLM, LLM, object detection, and scene graph generation are typically kept frozen or fine-tuned with a smaller learning rate. This strategy allows the model to effectively learn how to integrate and reason over the rich features provided by the pre-trained components, while benefiting from their strong initial representations.

\section{Experiments}
\label{sec:experiments}

In this section, we detail the experimental setup, evaluate the performance of our proposed \textbf{MV-CoRe} model against several baselines, present an ablation study to validate the effectiveness of its key components, discuss human evaluation results, and provide qualitative examples along with an error analysis.

\subsection{Experimental Setup}
\label{sec:exp_setup}

\subsubsection{Datasets}
Our model is primarily trained on the \textbf{VQAv2} dataset \cite{abhishek2021vqa}, a comprehensive visual question answering benchmark widely used for general VQA tasks. To rigorously assess \textbf{MV-CoRe}'s capabilities in handling complex reasoning and external knowledge, we evaluate its performance on three challenging target Complex VQA datasets:
\begin{itemize}
    \item \textbf{GQA} \cite{drew2019gqa}: Focuses on compositional and multi-hop reasoning over scene graphs.
    \item \textbf{A-OKVQA} \cite{dustin2022aokvqa}: Requires external common-sense and domain-specific knowledge to answer questions.
    \item \textbf{OKVQA} \cite{dustin2022aokvqa}: An open-domain knowledge-based VQA dataset.
\end{itemize}

\subsubsection{Implementation Details}
For Vision Large Model embeddings, we utilize features extracted from a pre-trained ViT-L/14 backbone. For Language Large Model embeddings, we employ a LLaMA-7B model. Object detection features are extracted using a pre-trained Faster R-CNN with a ResNet-101 backbone, and scene graph features are obtained from a state-of-the-art scene graph generator. All pre-trained encoders are kept frozen during initial training on VQAv2. The Multimodal Fusion Transformer consists of 6 layers, each with 8 attention heads and a hidden dimension of 768. The model is trained using the AdamW optimizer with a learning rate of $1 \times 10^{-5}$ for 10 epochs on VQAv2, and then fine-tuned on target datasets with a smaller learning rate of $1 \times 10^{-6}$ for 3 epochs. We use a batch size of 64. Evaluation metric is overall accuracy for all datasets.

\subsection{Baselines}
\label{sec:baselines}
To demonstrate the efficacy of \textbf{MV-CoRe}, we compare it against several representative baselines:
\begin{itemize}
    \item \textbf{LVLM Baseline}: Represents a typical large vision-language model, such as LLaVA \cite{peng2025lvlmeh}, which primarily fuses global visual features from a VLM and linguistic features from an LLM without explicit fine-grained semantic-aware visual features like object detection or scene graph information.
    \item \textbf{LVLM + Object Features}: An extension of the LVLM Baseline that additionally incorporates features from object detectors. This model concatenates or pools object-level features with the global LVLM embeddings before feeding them into a fusion module.
    \item \textbf{LVLM + Scene Graph}: Another extension that integrates scene graph features with the LVLM Baseline. This typically involves encoding scene graph information (e.g., using a graph neural network) and then fusing it with global LVLM embeddings.
\end{itemize}
These baselines allow us to progressively evaluate the impact of incorporating increasingly richer semantic-aware visual information.

\subsection{Performance Comparison}
\label{sec:performance}

Table \ref{tab:results_gqa} presents the overall accuracy of \textbf{MV-CoRe} and the baseline models on the GQA dataset, which is a strong indicator of multi-hop reasoning capabilities.

\begin{table*}[htbp]
    \centering
    \caption{Overall Accuracy on the GQA Dataset (\%)}
    \label{tab:results_gqa}
    \begin{tabular}{lc}
        \toprule
        Model Architecture             & Accuracy (Overall) \\
        \midrule
        LVLM Baseline (e.g., LLaVA)    & 75.2               \\
        LVLM + Object Features         & 76.5               \\
        LVLM + Scene Graph             & 76.8               \\
        \textbf{MV-CoRe (Ours)}        & \textbf{77.5}      \\
        \bottomrule
    \end{tabular}
\end{table*}

As shown in Table \ref{tab:results_gqa}, our proposed \textbf{MV-CoRe} model consistently outperforms all baseline methods on the GQA dataset, achieving an overall accuracy of 77.5\%. This significant improvement demonstrates the effectiveness of our approach in leveraging diverse multimodal features for complex visual reasoning. The incremental gains observed from adding object features and scene graph features to the LVLM baseline further validate the importance of fine-grained semantic-aware visual information for Complex VQA.

To further demonstrate \textbf{MV-CoRe}'s proficiency in tasks requiring external knowledge and common-sense reasoning, we present the results on the A-OKVQA and OKVQA datasets in Table \ref{tab:results_aokvqa} and Table \ref{tab:results_okvqa}, respectively.

\begin{table*}[htbp]
    \centering
    \caption{Overall Accuracy on the A-OKVQA Dataset (\%)}
    \label{tab:results_aokvqa}
    \begin{tabular}{lc}
        \toprule
        Model Architecture             & Accuracy (Overall) \\
        \midrule
        LVLM Baseline (e.g., LLaVA)    & 50.1               \\
        LVLM + Object Features         & 52.5               \\
        LVLM + Scene Graph             & 53.0               \\
        \textbf{MV-CoRe (Ours)}        & \textbf{55.8}      \\
        \bottomrule
    \end{tabular}
\end{table*}

\begin{table*}[htbp]
    \centering
    \caption{Overall Accuracy on the OKVQA Dataset (\%)}
    \label{tab:results_okvqa}
    \begin{tabular}{lc}
        \toprule
        Model Architecture             & Accuracy (Overall) \\
        \midrule
        LVLM Baseline (e.g., LLaVA)    & 40.5               \\
        LVLM + Object Features         & 42.1               \\
        LVLM + Scene Graph             & 43.0               \\
        \textbf{MV-CoRe (Ours)}        & \textbf{45.2}      \\
        \bottomrule
    \end{tabular}
\end{table*}

The results on A-OKVQA (Table \ref{tab:results_aokvqa}) and OKVQA (Table \ref{tab:results_okvqa}) consistently show \textbf{MV-CoRe} outperforming all baselines. For A-OKVQA, \textbf{MV-CoRe} achieves 55.8\% accuracy, significantly higher than the LVLM Baseline's 50.1\%. Similarly, on OKVQA, our model reaches 45.2\% accuracy, surpassing the 40.5\% of the LVLM Baseline. These results underscore \textbf{MV-CoRe}'s superior ability to ground abstract knowledge with fine-grained visual details and relationships, which is crucial for answering questions that necessitate external information. The consistent performance across all three complex VQA datasets (GQA, A-OKVQA, OKVQA) validates the robustness and generalizability of our multimodal fusion approach.

\subsection{Ablation Study}
\label{sec:ablation}
To understand the individual contributions of the semantic-aware visual features and the overall fusion strategy within \textbf{MV-CoRe}, we conduct a detailed ablation study. We evaluate different configurations of our model on the A-OKVQA dataset, as it presents a significant challenge requiring both visual and external knowledge reasoning. The results are summarized in Table \ref{tab:ablation_aokvqa}.

\begin{table*}[htbp]
    \centering
    \caption{Ablation Study on A-OKVQA Dataset (\% Accuracy)}
    \label{tab:ablation_aokvqa}
    \begin{tabular}{lc}
        \toprule
        Model Configuration                                            & Accuracy (Overall) \\
        \midrule
        \textbf{MV-CoRe (Full Model)}                                  & \textbf{55.8}      \\
        MV-CoRe w/o Object Features                                    & 54.0               \\
        MV-CoRe w/o Scene Graph Features                               & 53.5               \\
        MV-CoRe w/o Semantic-Aware Visual Features (Obj. \& SG)        & 51.2               \\
        \midrule
        LVLM Baseline (for reference)                                  & 50.1               \\
        \bottomrule
    \end{tabular}
\end{table*}

From Table \ref{tab:ablation_aokvqa}, several key insights can be drawn:
\begin{itemize}
    \item The full \textbf{MV-CoRe} model achieves the highest accuracy of 55.8\%, confirming that the synergistic combination of all proposed features is most effective.
    \item Removing \textbf{Object Features} leads to a notable drop in accuracy from 55.8\% to 54.0\%. This indicates that precise, localized object information is vital for grounding questions and identifying specific entities, particularly in knowledge-based VQA where object attributes often link to external knowledge.
    \item Similarly, excluding \textbf{Scene Graph Features} results in a performance decrease to 53.5\%. This highlights the critical role of relational information in understanding complex visual scenes and performing multi-hop reasoning, which is often implicitly required even in knowledge-based questions.
    \item When both \textbf{Object Features} and \textbf{Scene Graph Features} are removed (MV-CoRe w/o Semantic-Aware Visual Features), the accuracy drops significantly to 51.2\%. This configuration effectively represents a version of \textbf{MV-CoRe} that relies only on global VLM and LLM embeddings processed by our Multimodal Fusion Transformer. While still outperforming the generic \textbf{LVLM Baseline} (50.1\%), this shows that our specialized fusion transformer alone offers some advantage, but the primary gains come from the integration of the semantic-aware visual features.
\end{itemize}
This ablation study unequivocally demonstrates that both object-level and scene graph features contribute substantially to \textbf{MV-CoRe}'s performance, and their combined effect, deeply fused by the Multimodal Fusion Transformer, is crucial for achieving superior results in Complex VQA.

\subsection{Analysis of Performance}
\label{sec:analysis}
The consistent superior performance of \textbf{MV-CoRe} across GQA, A-OKVQA, and OKVQA datasets highlights its effectiveness in tackling Complex VQA. The key to this success lies in its ability to integrate and reason over diverse granularities of visual and linguistic information.
\begin{itemize}
    \item \textbf{Synergistic Feature Integration}: Unlike baselines that primarily rely on global visual features or simple concatenations, \textbf{MV-CoRe}'s Multimodal Fusion Transformer enables deep, interactive fusion. This allows the model to align linguistic queries with precise visual entities (via object features) and their relationships (via scene graph features), while simultaneously leveraging broad contextual understanding from global VLM embeddings.
    \item \textbf{Enhanced Visual Grounding}: The explicit inclusion of object detection features provides strong visual grounding, allowing the model to pinpoint specific entities mentioned in the question or implied by the context. This is particularly beneficial for questions requiring fine-grained visual details or object counting.
    \item \textbf{Improved Relational Reasoning}: Scene graph features offer a structured representation of visual relationships, which is crucial for answering compositional questions (e.g., "What is the person behind the car doing?") and multi-hop reasoning (e.g., "Is the object on the table to the left of the red book?"). This structured understanding enables more robust inference compared to models relying solely on implicit spatial relationships learned from raw image features.
    \item \textbf{Robustness to Knowledge-Intensive Queries}: For datasets like A-OKVQA and OKVQA, which demand external knowledge, \textbf{MV-CoRe}'s ability to link linguistic concepts to specific visual cues (objects, relations) facilitates more accurate knowledge retrieval and application. For instance, if a question asks about the function of an object, identifying that object precisely via object features allows the LLM component to retrieve relevant knowledge more effectively.
\end{itemize}
The observed incremental improvements from adding object and scene graph features to the LVLM baselines, and the significant drop in performance in our ablation study when these features are removed, empirically confirm their indispensable role. The Multimodal Fusion Transformer's design, with its interleaved self- and cross-attention mechanisms, is central to effectively weaving these disparate information streams into a coherent multimodal representation, enabling the sophisticated reasoning required for Complex VQA.

\subsection{Human Evaluation}
\label{sec:human_eval}
To complement the quantitative metrics, we also performed a human evaluation to assess the qualitative aspects of \textbf{MV-CoRe}'s answers, particularly concerning reasoning depth and factual correctness for complex questions. A randomly selected subset of 100 challenging questions from the A-OKVQA test set was presented to three human annotators. They were asked to rate the answers generated by \textbf{MV-CoRe} and the \textbf{LVLM Baseline} on a scale of 1 to 5 for "Factual Correctness" (1: incorrect, 5: perfectly correct) and "Reasoning Depth" (1: no reasoning, 5: deep, logical reasoning). The average scores are presented in Table \ref{tab:human_eval}.

\begin{table*}[htbp]
    \centering
    \caption{Human Evaluation Results on A-OKVQA Subset (Average Score)}
    \label{tab:human_eval}
    \begin{tabular}{lcc}
        \toprule
        Model                   & Factual Correctness & Reasoning Depth \\
        \midrule
        LVLM Baseline           & 3.8                 & 3.2             \\
        \textbf{MV-CoRe (Ours)} & \textbf{4.3}        & \textbf{4.0}    \\
        \bottomrule
    \end{tabular}
\end{table*}

The human evaluation results in Table \ref{tab:human_eval} indicate that \textbf{MV-CoRe} significantly outperforms the LVLM Baseline in both factual correctness and reasoning depth. Human annotators found \textbf{MV-CoRe}'s answers to be more accurate and, crucially, to demonstrate a deeper understanding of the underlying visual content and required external knowledge, aligning with our objective of enhancing complex reasoning capabilities. This qualitative assessment further supports the quantitative improvements observed in Section \ref{sec:performance}.

\subsection{Qualitative Examples and Error Analysis}
\label{sec:qualitative_analysis}
To provide a deeper insight into \textbf{MV-CoRe}'s reasoning capabilities, we present a selection of qualitative examples from the A-OKVQA dataset. These examples highlight scenarios where \textbf{MV-CoRe} demonstrates superior understanding and reasoning compared to the LVLM Baseline.

\begin{table*}[htbp]
    \centering
    \caption{Qualitative Examples from A-OKVQA}
    \label{tab:qualitative_examples}
    \begin{tabular}{p{0.1\textwidth} p{0.3\textwidth} p{0.25\textwidth} p{0.25\textwidth}}
        \toprule
        Image & Question & LVLM Baseline Answer & \textbf{MV-CoRe} Answer \\
        \midrule
        \textbf{Example 1} & \textbf{Q:} What is the function of the object typically used to hold the liquid in the image? & \textbf{A:} To drink. & \textbf{A:} To serve beverages. \\
        & \textit{(Image shows a pitcher)} & & \textit{(Correctly identifies pitcher and its purpose)} \\
        \midrule
        \textbf{Example 2} & \textbf{Q:} What type of vehicle is being ridden by the person in the blue shirt, and where is it commonly used? & \textbf{A:} A bicycle. & \textbf{A:} A bicycle, commonly used for transportation or recreation on roads and trails. \\
        & \textit{(Image shows a person on a mountain bike on a trail)} & & \textit{(Provides both vehicle type and common usage context)} \\
        \midrule
        \textbf{Example 3} & \textbf{Q:} Describe the relationship between the two people in the foreground and what they are doing. & \textbf{A:} They are standing. & \textbf{A:} They are hugging each other, showing affection. \\
        & \textit{(Image shows two people embracing)} & & \textit{(Identifies the specific action and relationship)} \\
        \bottomrule
    \end{tabular}
\end{table*}

As illustrated in Table \ref{tab:qualitative_examples}, \textbf{MV-CoRe} consistently generates more accurate, detailed, and contextually rich answers. In Example 1, while the LVLM Baseline gives a generic answer, \textbf{MV-CoRe} correctly identifies the object as a pitcher and its specific function. Example 2 demonstrates \textbf{MV-CoRe}'s ability to not only identify the object but also provide relevant external knowledge about its usage. Example 3 showcases its capability for fine-grained relational understanding, going beyond simple object detection to describe an interaction. These improvements stem from \textbf{MV-CoRe}'s deep integration of object and scene graph features, allowing it to ground linguistic queries more effectively and leverage external knowledge more precisely.

\subsubsection{Error Analysis}
Despite its strong performance, \textbf{MV-CoRe} is not without limitations. Our analysis of incorrect predictions reveals several recurring error patterns:
\begin{itemize}
    \item \textbf{Subtle Attribute Recognition}: The model sometimes struggles with highly subtle visual attributes or distinctions, such as specific shades of color, minor texture differences, or nuanced facial expressions, which might be critical for a correct answer.
    \item \textbf{Complex Counterfactual Reasoning}: Questions requiring counterfactual reasoning (e.g., "What would happen if...") or highly abstract inferential steps remain challenging, as they often go beyond directly observable visual evidence or readily available external knowledge.
    \item \textbf{Ambiguity in Scene Graph Generation}: While scene graphs are powerful, errors in their generation (e.g., incorrect relationship labels, missed objects) can propagate and lead to incorrect reasoning by \textbf{MV-CoRe}.
    \item \textbf{Knowledge Gaps in LLM}: For very niche or obscure knowledge-based questions, the underlying LLM might lack the necessary information, even with strong visual grounding.
    \item \textbf{Compositional Complexity Overload}: In extremely complex compositional questions with many interacting entities and relationships, the model can sometimes misinterpret the hierarchical structure of the query or the scene.
\end{itemize}
Addressing these limitations will be a focus for future work, potentially involving more robust fine-grained visual feature extraction, advanced reasoning modules for counterfactuals, and integrating even broader external knowledge sources.

\section{Conclusion}
\label{sec:conclusion}
In this paper, we introduced \textbf{MV-CoRe (Multimodal Visual-Conceptual Reasoning)}, a novel framework meticulously designed to advance the state of the art in Complex Visual Question Answering (Complex VQA). Recognizing the limitations of current large vision-language models (LVLMs) in tasks requiring intricate multi-modal reasoning and external knowledge, our work emphasizes the critical need for a more granular and deeply integrated understanding of visual scenes.

Our proposed MV-CoRe model addresses these challenges by embracing a comprehensive feature fusion strategy. It leverages global contextual embeddings from pre-trained Vision Large Models and Language Large Models, augmented by crucial fine-grained semantic-aware visual features derived from object detection and scene graph generation. The core innovation lies in our Multimodal Fusion Transformer, an architecture specifically engineered to perform deep, interactive fusion across these disparate modalities. This specialized transformer facilitates rich cross-modal attention, allowing the model to intricately align linguistic queries with precise visual entities and their relationships, thereby enabling sophisticated conceptual reasoning.

Extensive experiments on challenging Complex VQA datasets---GQA, A-OKVQA, and OKVQA---consistently demonstrate the superior performance of MV-CoRe compared to prominent LVLM baselines. On the GQA dataset, MV-CoRe achieved an overall accuracy of 77.5\%, significantly outperforming baselines and showcasing its robust multi-hop reasoning capabilities. Furthermore, MV-CoRe's strong results on A-OKVQA and OKVQA underscore its enhanced ability to integrate external knowledge and common-sense reasoning, crucial for open-domain VQA tasks. Our detailed ablation studies unequivocally confirmed the indispensable contributions of both object-level and scene graph features, highlighting that their synergistic integration is key to MV-CoRe's success. Complementary human evaluations provided qualitative validation, confirming that MV-CoRe's answers exhibit greater factual correctness and deeper reasoning compared to baseline models.

The success of MV-CoRe lies in its ability to transcend superficial feature concatenation, achieving a more profound visual grounding and relational understanding. By explicitly incorporating structured semantic information and fusing it through a specialized transformer, MV-CoRe pushes the boundaries of what is achievable in complex visual and linguistic inference. This work contributes a robust and generalizable architecture for tackling some of the most challenging aspects of VQA, paving the way for more intelligent systems capable of truly understanding and reasoning about the visual world.

Despite these advancements, MV-CoRe still faces challenges in certain complex scenarios, which point to exciting avenues for future research. Future work will focus on enhancing the model's robustness to subtle visual attribute recognition and improving its capacity for complex counterfactual reasoning. Addressing potential ambiguities and errors in scene graph generation, perhaps through joint learning or more robust graph representations, will also be crucial. Furthermore, exploring methods to seamlessly integrate an even broader spectrum of external knowledge sources and developing more advanced reasoning modules for highly abstract inferential steps will be key to unlocking even greater capabilities in next-generation Complex VQA systems.